\documentclass[11pt,twoside]{article}
\usepackage{informat}
\usepackage{epsfig}

\usepackage{graphicx}
\usepackage{algorithm}
\usepackage{algpseudocode}
\usepackage{amsmath}
\usepackage{multirow}

\begin{document}
  \title{Simulated Annealing with Levy Distribution for Fast Matrix Factorization-Based Collaborative Filtering}
   \author{Mostafa A. Shehata, Mohammad Nassef and Amr A. Badr \\
Department of Computer Science, Faculty of Computers and Information Technology, Cairo University, 5 Dr. Ahmed Zewail Street, Orman, Giza 12613, Egypt.\\
      mostafaashraf413@yahoo.com\\
      m.nassef@fci-cu.edu.eg\\
      amr.badr@fci-cu.edu.eg}
%   \titleodd{Title of the\ldots}
%   \authoreven{N. SurnameofAuthor1 et al.}
 
 \received{June 24, 2013}

  \abstract{Matrix factorization is one of the best approaches for collaborative filtering, because of its high accuracy in presenting users and items latent factors.
The main disadvantages of matrix factorization are its complexity, and being very hard to be parallelized, specially with very large matrices. In this paper, we introduce a new method for collaborative filtering based on Matrix Factorization by combining simulated annealing with levy distribution. By using this method, good solutions are achieved in acceptable time with low computations, compared to other methods like stochastic gradient descent, alternating least squares, and weighted non-negative matrix factorization.}
%   \abstractSi{}

  \maketitle

\section{INTRODUCTION}
\label{intro}
% \paragraph{Recommender Systems}
The objective of Recommender Systems is to recommend new products or items for users based on their history \cite{recommender_handbook}. There are two major approaches to create a Recommender System. The first one is the \textit{Content Filtering} or (\textit{Content Based}). This approach tries to create a profile for each user and item, and then tries to match these profiles\cite{yahooATT}. The second approach is the \textit{Collaborative Filtering}. It uses the rating history of users and items, and creates a large sparse matrix called Rating Matrix. This matrix usually contains ratings from 1 to 5. The 0's are for the incomplete ratings. The objective of the \textit{Collaborative Filtering} is to predict these missing ratings. One of the most successful method for \textit{Collaborative Filtering} is the \textit{Latent Factor Model}\cite{yahooATT}. This method tries to learn the latent features of each user and item in a fixed number of dimensions. Then represent each of them in a latent feature vector, that can be used to predict the incomplete ratings, or measure the similarity.

% \subsection*{\textbf{Problem Statement}}
% \label{prob-stat}
\textit{Matrix Factorization} is one of the best techniques used for \textit{Latent Factor Model}. The basic idea is to construct the \textit{low-dimensional} matrices to approximate the original rating matrix \cite{yahooATT} \cite{nmf-cf} \cite{nmf} \cite{bayesian_nmf_cf},
\begin{equation}
R \approx U \cdotp I
\end{equation}
where $R_{M,N}$ is the rating matrix, $U_{M,K}$ is the users matrix, $I_{K,N}$ is the items matrix. $M$ and $N$ are the number of users and items respectively, $K$ is the number of latent feature that represent each user and item. Where $K \ll \min (M,N)$.
Row $m$ in matrix $U$ represents user number $m$ in the rating matrix, whereas column $n$ in matrix $I$ represents item number $n$ in the rating matrix. So, in the rating matrix, the rating of user $m$ for item $n$ can be calculated by the \textit{dot product} of row $m$ of matrix $U$ by the column $n$ of matrix $I$.
\begin{equation}
r_{m,n} \approx u_m \cdotp i^T_n
\end{equation}
The rating matrix is very sparse, because it contains a few users ratings. The objective of this paper is to use the known ratings to construct the \textit{low-rank} matrices, to predict the unknown or incomplete ratings.

One of the most common evaluation metrics for \textit{Collaborative Filtering} is \textit{RMSE} (root mean squared error). We calculate \textit{RMSE} only for the known rating using the following equation:
\begin{equation}
RMSE = \sqrt[2]{(\sum_{m,n \in KR} (r_{m,n} - (u_m \cdotp i^T_n))^2)/ |KR|}
\end{equation}
where KR is the list of known ratings.

% \subsection*{\textbf{Related Work}}
% \label{rw}
There is a lot of work done on Matrix Factorization and Collaborative Filtering. Here we discus three of the most popular methods. 
\subsubsection*{Stochastic Gradient Descent (SGD)}
SGD is one of the popular Matrix Factorization methods \cite{yahooATT}. The idea is to minimize the following cost equation:
\begin{align}
\label{SGD_min_fun}
\begin{split}
 \min_{u^*,i^*,b^*} \sum_{(m,n)\in KR} (r_{m,n} - \mu - b_m - b_n - u_m \cdot i^T_n)^2 +  \\ \lambda (\|u_m\|^2 + \|i_n\|^2 + b_m^2 + b_n^2)
\end{split}
\end{align}
where $\lambda$ is a regularization term, $\mu$ is the overall average rating, $b_m$ and $b_n$ are the user and item bias respectively.
\begin{equation}
\label{SGD_error_fun}
e_{m,n} = r_{m,n} - \mu - b_m - b_n - u_m \cdot i^T_n
\end{equation}
To minimize the squared-error equation \eqref{SGD_min_fun}, the algorithm iterates over all ratings in the training set. Then, it computes the associated prediction error in equation(\ref{SGD_error_fun}). Next, the error value is used to compute the gradient. The algorithm finally uses the gradient to update user bias, item bias, user matrix $U$, and item matrix $I$.
\begin{equation}
\label{SGD_user_bias_update}
b_m = b_m + \gamma (e_{m,n} - \lambda b_m)
\end{equation}
\begin{equation}
\label{SGD_item_bias_update}
b_n = b_n + \gamma (e_{m,n} - \lambda b_n)
\end{equation}
\begin{equation}
\label{SGD_usr_update_fun}
u_m = u_m + \gamma (e_{m,n} \cdot i_n - \lambda \cdot u_m)
\end{equation}
\begin{equation}
\label{SGD_item_update_fun}
i_n = i_n + \gamma (e_{m,n} \cdot u_m - \lambda \cdot i_n)
\end{equation}
where $\gamma$ is the learning rate. The learning rate determines the moving speed towards the optimal solution. If $\gamma$ is very large, we might skip the optimal solution. If it is too small, we may need too many iterations to reach the optimal solution. So using an appropriate $\gamma$ is very important.

\subsubsection*{Alternating Least Squares (ALS)}
ALS is very good for parallelization \cite{als}. When we have large data, and need to distribute the computations over cluster of nodes. ALS objective is to minimize the following equation:
\begin{equation}
\label{ALS_min_fun}
 \min_{u^*,i^*,b^*} \sum_{(m,n)\in KR} (r_{m,n} - u_m \cdot i^T_n)^2 + \lambda (\|u_m\|^2 + \|i_n\|^2)
\end{equation}
It is the same like equation \eqref{SGD_min_fun}, but without the bias terms. the basic idea can be summarized as follows:
\begin{enumerate}
\item Initialize $U$, and $I$ matrices.
\item Fix $I$, solve for $U$ by minimizing equation \eqref{ALS_min_fun}.
\item Fix $U$, solve for $I$ by minimizing equation \eqref{ALS_min_fun}.
\item Repeat the previous two steps until converging or reaching the max iteration.
\end{enumerate}
to solve the user $U$ and item $I$ matrices we use the following two equations respectively:
\begin{equation}
u^T_m = (r_m \cdot I) \cdot (I^T \cdot I + \lambda Eye)^{-1}
\end{equation}
\begin{equation}
i^T_n = (r^T_n \cdot U) \cdot (U^T \cdot U + \lambda Eye)^{-1}
\end{equation}
where $Eye$ is the Identity matrix.

\subsubsection*{Weighted Non-Negative Matrix Factorization (WNMF)}
Here we present a special type of Matrix Factorization called \textit{Non-Negative Matrix Factorization} (NMF). The only difference is the non-negativity constraint for the input matrix, and the low rank matrices as well. The problem can be formulated as an optimization problem:
\begin{equation}
\label{nmf_eq}
\begin{aligned}
&\min_{A,H} &&\|V-A \cdot H\|^2\\
&\text{subject to} &&A, H \geq 0
\end{aligned}
\end{equation}
where $V$ is the original matrix, $W and H$ are the two factorized matrices. 
One of the most simplest methods for NMF is \textit{Multiplicative Update Rules} \cite{nmf}. It is a good compromise between speed and ease of implementation. So NMF objective equation \eqref{nmf_eq} can be optimized using the following update rules:
\begin{equation}
A^{(t+1)} = A^{(t)} \frac{V \cdot H^T}{A \cdot H \cdot H^T}
\end{equation}
\begin{equation}
H^{(t+1)} = H^{(t)} \frac{A^T \cdot V}{A^T \cdot A \cdot H}
\end{equation}
The original version of \textit{Multiplicative Update Rules} will not fit in our problem. The two rules will not be able to differentiate between the true ratings, and the incomplete ratings. So we need to modify the original rules, to be able to learn from the true ratings, then predict the incomplete. In \cite{wnmf} they could modify the original \textit{Multiplicative Update Rules} to be able to do Incomplete Matrix Factorization. This method called \textit{Weighted Non-Negative Matrix Factorization (WNMF)}. So now the new objective function is:
\begin{equation}
\label{WNMF_min_fun}
 \min_{u^*,i^*} \sum_{(m,n)\in KR} (r_{m,n} - u_m \cdot i^T_n)^2 
\end{equation}
It is similar to equations \eqref{SGD_min_fun}, \eqref{ALS_min_fun}, but without the bias $b$ or regularization $\lambda$. Now we can optimize function \eqref{WNMF_min_fun} using the following two rules:
\begin{equation}
U^{(t+1)} = U^{(t)} \frac{(W*R) \cdot I^T}{(W*(U \cdot I)) \cdot I^T}
\end{equation}
\begin{equation}
I^{(t+1)} = I^{(t)} \frac{U^T \cdot (W*R)}{U^T \cdot (W *(U \cdot I))}
\end{equation}
where $W_{M,N}$ is a matrix which its elements are equal to $1$ if the corresponding entry in $R$ is known rating, and $0$ otherwise. ($*$) denotes to the element wise multiplication.

In this paper we focus on efficiency more than effectiveness. We assume that we have a very large data, and limited time. So we need an acceptable solution in reasonable time. So we chose the Metaheuristic algorithms for this problem, because of its ability to scape from the local optimal, and reaching good solutions in reasonable time. We used Simulated Annealing algorithm, with Levy Flight as a random walk operator. 

The rest of the paper is organized as follows: in section (\ref{pre_sec}) we briefly describe the prerequisite topics that are needed before going through the proposed method. In section (\ref{solution_sec}) we describe our proposed method. In section (\ref{exp_sec}) we discuss the experimental results, effect of each parameter, and compare our proposed method against others. Finally section (\ref{conc_sec}) concludes the paper.

\section{PRELIMINARIES}
\label{pre_sec}
\subsection{\textbf{Metaheuristic Optimization}}
\label{meta-sec}
There are two types of optimization algorithms, \textit{Deterministic} and \textit{Stochastic}. Deterministic algorithms usually focus on optimal solution, like Simplex method in linear programming, some of these algorithms use the derivative of the objective function, these algorithms are called \textit{gradient based algorithms}.

In \textit{Stochastic Optimization} we will talk about \textit{Metaheuristic Algorithms}, we can divide \textit{Metaheuristic} into two parts, META and HEURISTIC, META means "beyond" or "higher level", and HEURISTIC means "to find" or "to discover by trial and error". This type of algorithms depends on randomization and local search to find the optimal solution iteratively, whereas each iteration tries to improve the current solutions from previous iteration. Also \textit{Metaheuristic} doesn't guarantee the optimal solution, but it gives good quality solutions in a reasonable time. \textit{Metaheuristic} achieves its goal by making a good balance between two major components, intensification and diversification. Intensification is to search for a better solution within the local area of the current solution. Diversification is to use the randomization to escape from the local optimum, and explore all the search space \cite{xin_book}.

There are many types of \textit{Metaheuristic} algorithms, like single solution, or population based, in this paper we use \textit{Single Solution}.

\subsection{\textbf{Levy Distribution and Random Walk}}
We presented randomization techniques for exploring the search space (\textit{Diversification}), local search for optimizing the current solution, and searching within the local area of it (\textit{Intensification}). In \eqref{rweq} $x^t$ is the current solution state, $s$ is a new step or random number drawn from a probability distribution, we add $s$ to $x$ to move it from state $t$ to $t+1$.
\begin{equation}
\label{rweq}
x^{t+1} = x^t + s
\end{equation}
\textit{Levy Flights} are a random walk that their steps are drawn from \textit{Levy Distribution}. \textit{Mantegna} algorithm is the best and easiest way to generate random numbers from \textit{Levy Distribution} \cite{xin_book}\cite{mantegna}, so the random walk can be achieved  using the following equations:
\begin{equation}
\label{rw_equation}
x^{t+1}_i = x^t_i + \alpha L(s,\lambda),
\end{equation}
Where $\alpha$ is the step size.
\begin{equation}
L(s,\lambda) = \frac{\lambda \Gamma \sin(\pi \lambda / 2)}{\pi} \frac{1}{s^{1+\lambda}},  
\end{equation}
\begin{equation}
s = \frac{U}{|V|^{1/\alpha}}
\end{equation}
\begin{equation}
U ~ N(0,\sigma^2), V ~ N(0,1)
\end{equation}
Where $N$ is a Gaussian normal distribution
\begin{equation}
\sigma^2 = \bigg[\frac{\Gamma(1+\lambda)}{\lambda \Gamma((1+\lambda)/2)}.\frac{\sin(\pi \lambda/2)}{2^{(\lambda - 1)/2}}\bigg]^{1/\lambda} 
\end{equation}

\subsection{\textbf{Simulated Annealing}}
SA is one of the most popular metaheuristic algorithms. It simulates the annealing process for solids by cooling to reach the crystal state. Reaching the crystal state is like reaching the global optimum in optimization. It is a single solution algorithm. The basic idea is to perform a random walk, but with some probability called \textit{Transition Probability} that may accept new solutions that do not improve the objective function, see equation \eqref{transProb}. Accepting bad solutions with \textit{Transition Probability} gives more exploration for the search space (\textit{Diversification}). \textit{Transition Probability} decreases gradually during the iterations to decrease the \textit{Diversification} and increase the \textit{Intensification}. This means that the algorithm will end up with accepting only better solutions \cite{xin_book} \cite{annealing}.
\begin{equation}
\label{transProb}
p = \exp \bigg[ - \frac{\Delta f}{T} \bigg] > r
\end{equation}
In \eqref{transProb} $f$ is the difference between the two evaluation function values of the current solution and the new one. $T$ is the current temperature which is decreased iteratively by the cooling rate. $r$ is a random number. So the algorithm accepts bad solution if the \textit{Transition Probability} $p$ is greater than $r$.

One of the common cooling schedules is linear cooling schedule, in \eqref{coolEq}
\begin{equation}
\label{coolEq}
T = T_0 - \beta t
\end{equation}
$T_0$ is the initial temperature, $\beta$ is the cooling rate, t is the pseudo time for iterations. The following pseudo code demonstrate the basic implementation of Simulated Annealing.

\begin{algorithm}
\caption{Simulated Annealing}
\begin{algorithmic}[1]
\State Objective function $f(x), x = (x 1 , ..., x d )^T$
\State Initialize the initial temperature $T_0$ and initial guess $x (0)$
\State Set the final temperature $T_f$ and the max number of iterations $N$
\State Define the cooling schedule $T \mapsto \alpha T, (0< \alpha <1)$

\While{$( T > T_f$ and $t < N )$}
\State Drawn $\epsilon$ from a Gaussian distribution
\State Move randomly to a new location: $x_{t+1} = x_t+ \epsilon $(random walk)
\State Calculate $\Delta f = f_{t+1} (x_{t+1}) - f_t(x_t)$
\State Accept the new solution if better
\If {not improved}
\State Generate a random number $r$
\State Accept if $p = \exp [− \Delta f /T ] > r$
\EndIf
\State Update the best $x\star$ and $f\star$
\State $t = t +1$
\EndWhile
\end{algorithmic}
\end{algorithm}

\section{THE PROPOSED METHOD}
\label{solution_sec}
In this section we introduce our new method for solving Matrix Factorization. In our method we use \textit{Simulated Annealing} based on \textit{Levy Flight} as a random walk, instead of \textit{Gaussian distribution}. See section.(\ref{meta-sec}). We called it \textit{SA-Levy}. We choose \textit{Simulated Annealing} because its low computations, as it is a single solution \textit{Metaheuristic} algorithm. So it will be very fast compared to other population based \textit{Metaheuristic} algorithms. Also compared to the state of art methods like \textit{ALS}, and \textit{WNMF} \textit{Simulated Annealing} is much faster, because it needs only one matrix multiplication per iteration. Regarding \textit{SGD}, \textit{Simulated Annealing} is easier to be parallelized.

In \textit{Simulated Annealing} we just need to represent the solution, and implement the evaluation function to compare the current solution against others. We use \textit{RMSE} as an evaluation function.

\begin{figure}
  \includegraphics[width=\linewidth]{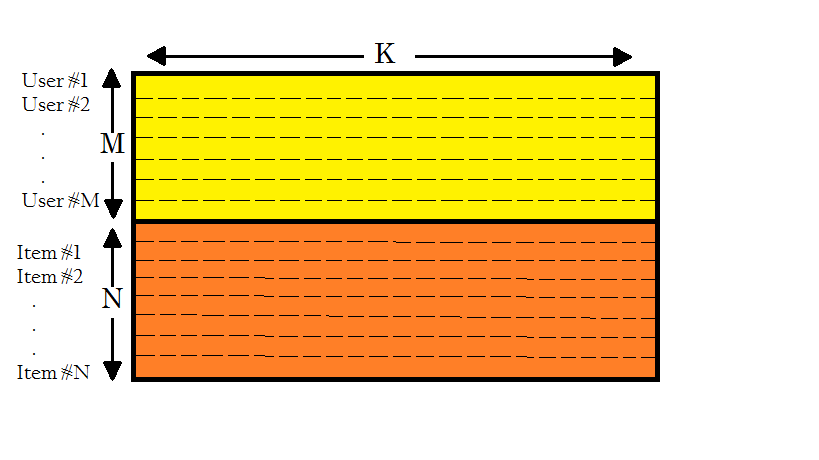}
  \caption{Representing $U$ matrix and $I$ matrix in one matrix, the number of rows is equal $M+N$, and the number of columns is equal $K$.}
  \label{sol_rep_fig}
\end{figure}

\begin{table*}
\caption{Shows the effect of the number of iteration on RMSE}
\label{iteration_table}       % Give a unique label
\vspace{2mm}

\centering
\begin{tabular}{|c|cccccc|}
\hline
%\noalign{\smallskip}
Iterations & 5 & \textbf{10} & 25 & 50 & 100 & 200  \\
%\noalign{\smallskip}
\hline
%\noalign{\smallskip}
RMSE & 1.112 & \textbf{1.118} & 1.118 & 1.118 & 1.118 & 1.118 \\
%\noalign{\smallskip}
\hline
\end{tabular}
\end{table*}

Figure \ref{sol_rep_fig} shows the representation of the solution, we put the two matrices users and items in one matrix to simplify the solution and the calculation \cite{genetic_sol}, the number of rows is equal the number of users $M$ plus the number of items $N$, and the number of columns is equal the number of latent features $K$.

\section{EXPERIMENTAL RESULTS}
\label{exp_sec}
%%%%%%%%%%%%%%%%%%%% Experimental tables

\begin{table}
\begin{small}
\caption{Shows the effect of number of latent features on RMSE}
\label{latent_features_table}
\vspace{2mm}

\begin{tabular}{|c|ccccc|}
\hline
%\noalign{\smallskip}
Latent  & \multirow{2}{*}{10} & \multirow{2}{*}{\textbf{20}} & \multirow{2}{*}{30} & \multirow{2}{*}{40} & \multirow{2}{*}{50}  \\
Features &&&&&\\
%\noalign{\smallskip}
\hline
%\noalign{\smallskip}
RMSE & 1.120 & \textbf{1.118} & 1.118 & 1.118 & 1.118 \\
%\noalign{\smallskip}
\hline
\end{tabular}
\end{small}
\end{table}

\begin{table}[!t]
\caption{Shows the effect of step size of Levy flight on RMSE}
\label{step_size_table}
\vspace{2mm}

\centering
\begin{tabular}{|c|ccc|}
\hline%\noalign{\smallskip}
Step Size & 0.1 & \textbf{0.01} & 0.001 \\
%\noalign{\smallskip}
\hline
%\noalign{\smallskip}
RMSE & 1.119 & \textbf{1.118} & 1.145 \\
%\noalign{\smallskip}
\hline
\end{tabular}
\end{table}

\begin{table}
\caption{Compares Levy against Gaussian distribution as a random walk for Simulated Annealing}
\label{levyVSnormal}
\vspace{2mm}

\centering
\begin{tabular}{|c|cc|}
\hline
% \noalign{\smallskip}
Distribution & \textbf{Levy} & Gaussian \\
% \noalign{\smallskip}
\hline
% \noalign{\smallskip}
RMSE & \textbf{1.118} & 1.168 \\
% \noalign{\smallskip}
\hline
\end{tabular}
\end{table}

\begin{table}
\caption{Compares SA-Levy with other methods (SGD, WNMF, and ALS) }
\label{SA-levyVSothers}
\vspace{2mm}

\begin{tabular}{|c|cccc|}
\hline
% \noalign{\smallskip}
System & SA-Levy & SGD & WNMF & ALS \\
% \noalign{\smallskip}
\hline
% \noalign{\smallskip}
RMSE & 1.118 & 0.871 & 0.943 & 1.007 \\
% \noalign{\smallskip}
\hline
\end{tabular}
\end{table}
%%%%%%%%%%%%%%%%%%%%%%%%%%%%%%%%%%%%%
In this section we show the effect of the parameters on the \textit{RMSE} results, we uses MovieLens 1M dataset \cite{1m_movielens} in our experiments, 80\% of the dataset is used for training and 20\% for testing.

Tables \ref{iteration_table} \ref{latent_features_table} \ref{step_size_table} show how \textit{RMSE} can be affected by the number of iteration, number of latent features, and step size. see equation (\ref{rw_equation}). In table \ref{iteration_table} we can see that good \textit{RMSE} can be achieved by few number of iterations, so there is no need for many iteration to converge. In table \ref{latent_features_table} we can see that best \textit{RMSE} can be achieved starting from 20 latent features. In table \ref{step_size_table} we found that the best value for the step size is 0.01. We can say that the step size is the most important parameter in our method. It manages the balance between Intensification and Diversification, see section (\ref{meta-sec}). Small values of step size give more Intensification, and and large values give more Diversification.

Table \ref{levyVSnormal} shows the difference between using Gaussian distribution and Levy distribution as a random walk. Levy distribution outperform Gaussian because of it's ability to escape from the local minimum \cite{xin_book}\cite{mantegna}.

Table \ref{SA-levyVSothers} shows that SA-Levy can be outperformed by other methods in terms of effectiveness. But SA-Levy can outperform all other methods in terms efficiency, because of its low computations, where it needs only one matrix multiplication in each iteration. Unlike WNMF or ALS which need many matrix multiplications or calculating matrix inversion in each iteration. Also it is much easier than SGD to be parallelized because it doesn't need huge amount of data to be shuffled between the cluster nodes. So SA-Levy can be a good choice if we have limited time or resources and large amount of data.

\subsection*{\textbf{Choice of Parameters}}
We conducted these experiments using \textbf{Simanneal}. It is a python module for simulated annealing optimization \footnote{https://github.com/perrygeo/simanneal}, also the project source code can be found here \footnote{https://github.com/mostafaashraf413/MF\_SA\_Levy}. Based on the MovieLens 1M dataset \cite{1m_movielens} we found that the best parameters are 10 Iteration, 20 latent features, 0.01 step size. For the temperature parameter we found that the best values for  maximum and minimum temperature are 25000 and 2.5 respectively. To focus more on Diversification at the beginning then decrease it gradually to increase the Intensification.

\section{CONCLUSION AND FUTURE WORK}
\label{conc_sec}
We presented in this work a new method for matrix factorization based Collaborative filtering. We achieved a significant improvement in Simulated Annealing, by using Levy distribution as a random walk, instead of Gaussian distribution. We expect this contribution could fit in many optimization problems, not only matrix factorization. We think that SA-Levy is a good choice for complex matrix factorization problems. When we have a very large data, and limited time for computation. We expect that SA-Levy can be easily implemented on any distributed system, that has basic linear algebra operations, like \textit{Apache Spark}\footnote{https://spark.apache.org/}, and \textit{Hadoop}\footnote{http://hadoop.apache.org/}.

% bibliography style is not predefined; authors can apply their own,
% but it should be sent to matjaz.gams@ijs.si with the accepted paper

\end{document}